%% file: root.tex
\documentclass[letterpaper, 10 pt, conference]{ieeeconf}

\IEEEoverridecommandlockouts                              % This command is only needed if 
\usepackage{times} % assumes new font selection scheme installed
\usepackage{amsmath} % assumes amsmath package installed
\usepackage{amssymb}  % assumes amsmath package installed
\usepackage{color}
\usepackage{booktabs}
\usepackage{graphicx}
\usepackage{subfigure}
\usepackage{multirow,booktabs}
\usepackage{stfloats} %fig location
\usepackage{fancyhdr}
\usepackage{footnote}
\usepackage{float}
\usepackage{pifont}
\usepackage[colorlinks, linkcolor=red, citecolor=blue]{hyperref}

\title{\LARGE \bf PanopticRecon: Leverage Open-vocabulary Instance Segmentation\\ for Zero-shot Panoptic Reconstruction}

\author{Xuan Yu$^{1}$, Yili Liu$^{1}$, Chenrui Han$^{1}$, Sitong Mao$^{2}$, Shunbo Zhou$^{2}$, Rong Xiong$^{1}$, Yiyi Liao$^{1}$, Yue Wang$^{1}$  %and Bernard D. Researcher$^{2}$% <-this % stops a space
\thanks{$^{1}$Xuan Yu, Yili Liu, Chenrui Han, Rong Xiong, Yiyi Liao and Yue Wang are with Zhejiang University, Hangzhou, Zhejiang, China. Yue Wang is the corresponding author {\tt\footnotesize wangyue@iipc.zju.edu.cn}}%
\thanks{$^{2} $Sitong Mao and Shunbo Zhou are with Huawei Cloud Computing Technologies Co., Ltd., Shenzhen, China.}}

\begin{document}

\maketitle
% \thispagestyle{empty}
% \pagestyle{empty}

\input{sec0_abstract}

\input{sec1_intro}

\input{sec2_related}
\input{sec3_method}

\input{sec4_exp}
\input{sec5_conclusion}
\input{bibliography}
% \input{appendix}


%%%%%%%%%%%%%%%%%%%%%%%%%%%%%%%%%%%%%%%%%%%%%%%%%%%%%%%%%%%%%%%%%%%%%%%%%%%%%%%%

%%%%%%%%%%%%%%%%%%%%%%%%%%%%%%%%%%%%%%%%%%%%%%%%%%%%%%%%%%%%%%%%%%%%%%%%%%%%%%%%



% \section*{ACKNOWLEDGMENT}

% The preferred spelling of the word ÒacknowledgmentÓ in America is without an ÒeÓ after the ÒgÓ. Avoid the stilted expression, ÒOne of us (R. B. G.) thanks . . .Ó  Instead, try ÒR. B. G. thanksÓ. Put sponsor acknowledgments in the unnumbered footnote on the first page.

\end{document}

%% file: sec0_abstract.tex
%%%%%%%%%%%%%%%%%%%%%%%%%%%%%%%%%%%%%%%%%%%%%%%%%%%%%%%%%%%%%%%%%%%%%%%%%%%%%%%%
\begin{abstract}
Panoptic reconstruction is a challenging task in 3D scene understanding. However, most existing methods heavily rely on pre-trained semantic segmentation models and known 3D object bounding boxes for 3D panoptic segmentation, which is not available for in-the-wild scenes. In this paper, we propose a novel zero-shot panoptic reconstruction method from RGB-D images of scenes. For zero-shot segmentation, we leverage open-vocabulary instance segmentation, but it has to face \textit{partial labeling} and \textit{instance association} challenges. We tackle both challenges by propagating partial labels with the aid of dense generalized features and building a 3D instance graph for associating 2D instance IDs. Specifically, we exploit partial labels to learn a classifier for generalized semantic features to provide complete labels for scenes with dense distilled features. Moreover, we formulate instance association as a 3D instance graph segmentation problem, allowing us to fully utilize the scene geometry prior and all 2D instance masks to infer global unique pseudo 3D instance ID. Our method outperforms state-of-the-art methods on the indoor dataset ScanNet V2 and the outdoor dataset KITTI-360, demonstrating the effectiveness of our graph segmentation method and reconstruction network.

\end{abstract}

%Concretely, we distill the semantic VLM feature into the neural implicit representation to propagate labels for unknown pixels. We also build pseudo 3D instance ID generated by inference on a graph, which cuts the edge to maximize the 3D-2D instance mask consistency. Furthermore, we propose a method to leverage the instance association to correct the per-frame semantic labels.
% Unlike previous methods, our approach leverages the generalization of the 2D foundation model to a greater extent and achieves an efficient 3D segmentation system by combining distillation features with a nonlinear classifier. \yw{??}
%%%%%%%%%%%%%%%%%%%%%%%%%%%%%%%%%%%%%%%%%%%%%%%%%%%%%%%%%%%%%%%%%%%%%%%%%%%%%%%%

%% file: sec1_intro.tex
\section{Introduction}

Panoptic reconstruction that builds the 3D geometry of the scene labeled with semantics and instance, is an important task in 3D scene understanding. The representation is ideal for localization and planning of autonomous robots, as well as interaction with human users, thus drawing the attention of the community.

Pioneering methods aim at semantic reconstruction~\cite{suma++} design rules for integration of scans labeled by semantic segmentation~\cite{mask2former}. One weakness of these methods is the separate optimization of geometry and semantics. Recent years witnesses great success in implicit neural representation~\cite{mildenhall2021nerf, wang2021neus}, which allows for end-to-end semantic reconstruction~\cite{semanticnerf,nesf,nfatlas}. Empowered by foundation visual language model (VLM)~\cite{lseg, oquab2023dinov2}, the open-vocabulary semantic cues are further employed for zero-shot semantic reconstruction to eliminate human annotation and segmentation network training for unseen class~\cite{Peng2023OpenScene,nivlff}.

PNF~\cite{pnf} and panoptic NeRF~\cite{panopticnerf} pioneeringly extend the semantic reconstruction method to panoptic reconstruction by replacing the semantic labels with 2D panoptic labels and 3D instance bounding box, and further assuming the known data association between unique 3D instance ID and per-frame 2D instance ID. For zero-shot panoptic reconstruction, since there is no open-vocabulary panoptic segmentation VLM, PVLFF~\cite{Chen2024PVLFF} proposes to distill features of two VLMs for open-vocabulary semantic segmentation (LSeg~\cite{lseg}) and instance segmentation (SAM~\cite{sam}). Since SAM is not able to achieve object-level segmentation, PVLFF struggles to segment semantically aligned object-level instances.

%respectively. When instance inheriting the label of semantic segmentation in standard panoptic segmentation\yl{didn't get}, the results of two independent VLMs lose consistency\yl{the prediction of PVLFF loses consistency or the pseudo-gt loses consistency?}, i.e., the predicted instance mask may have misaligned semantic label.

\begin{figure}[!t]
\centering
\includegraphics[width=\linewidth,keepaspectratio]{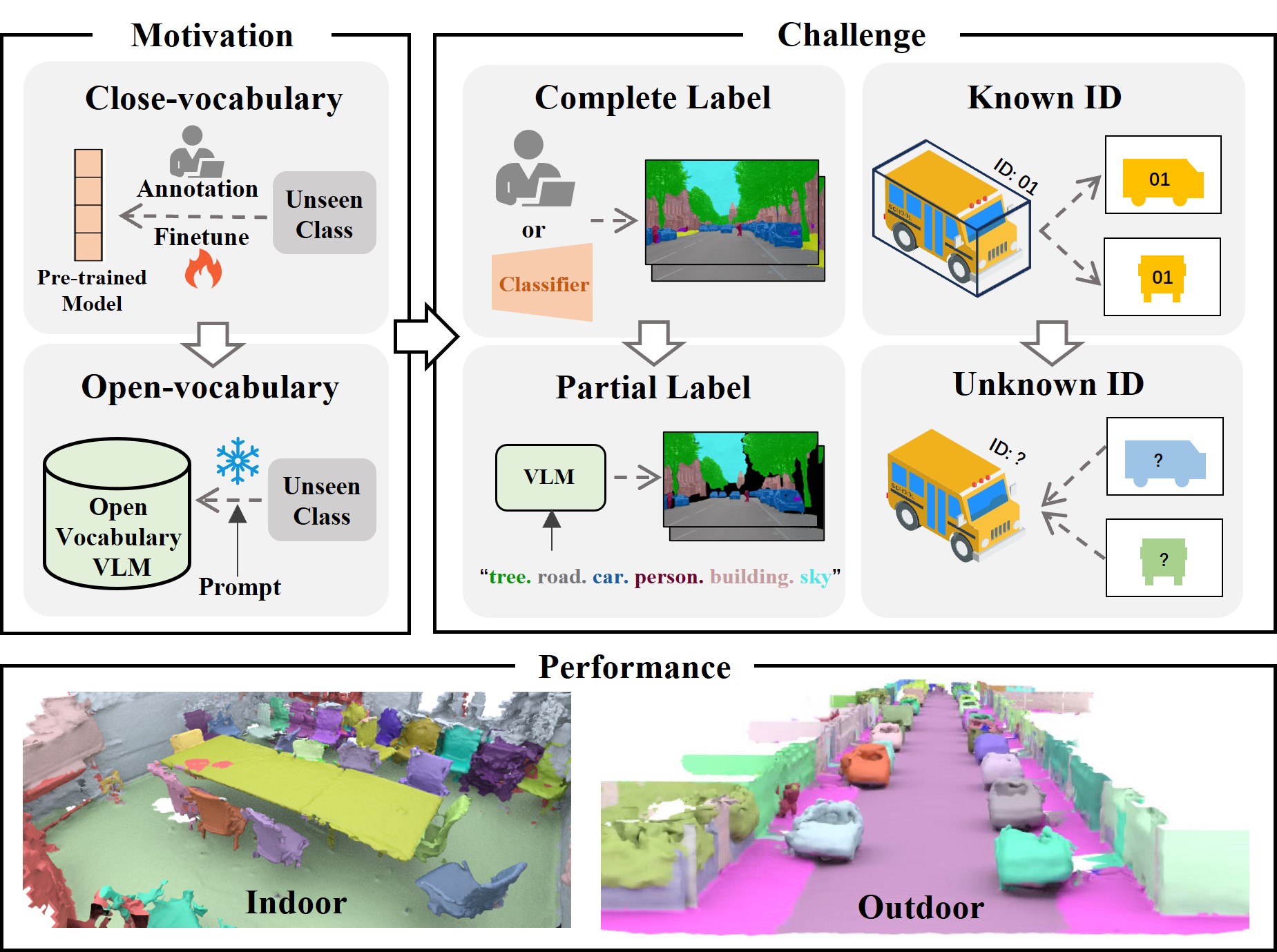}\\
\vspace{-0.3cm}
\caption{\label{Fig_main} Zero-shot panoptic reconstruction by leveraging open-vocabulary instance segmentation faces two challenges: 1) 2D semantic labels provided by text prompt based VLMs are not complete. 2) No object-level instance 3D pseudo ID makes 2D instance ID inconsistent. We supplement blank pixel labels with distilled DINOv2 features and establish a graph to infer 3D instance pseudo IDs.}
\vspace{-0.3cm}
\end{figure}

\begin{figure*}[!t]
\centering
\includegraphics[width=\linewidth,keepaspectratio]{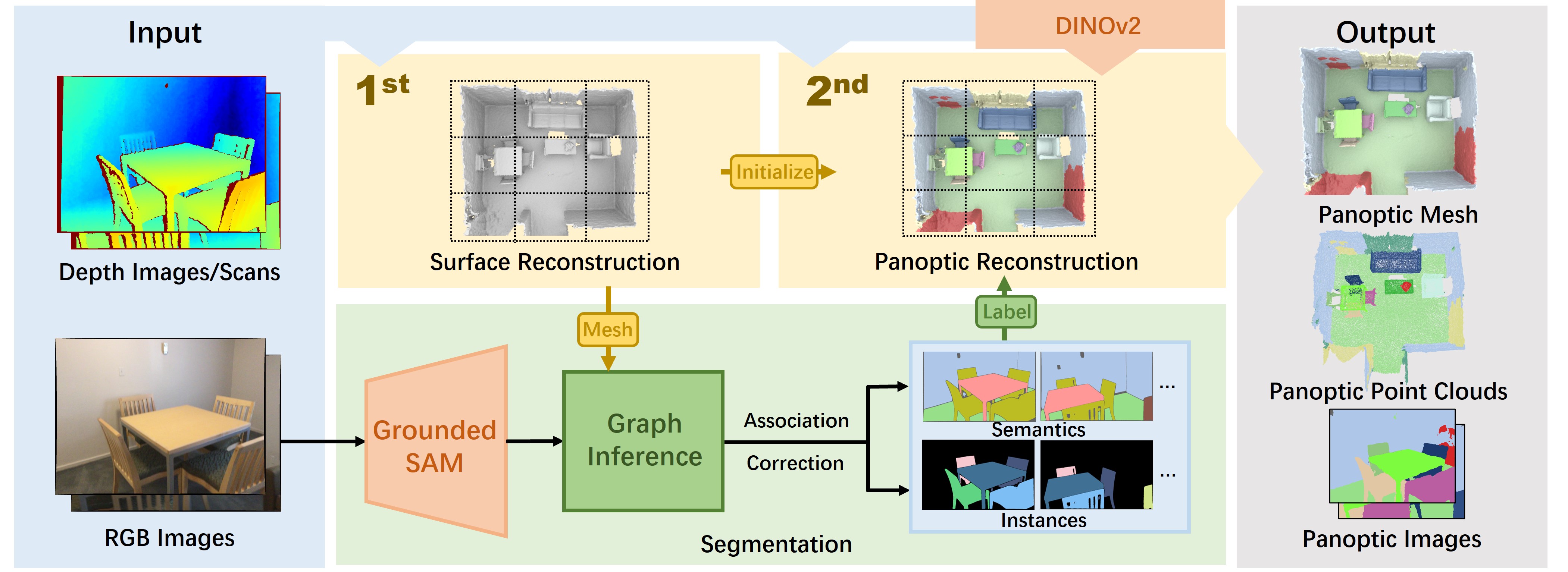}\\
\vspace{-0.4cm}
\caption{\label{Fig_pipeline}PanopticRecon consists of a reconstruction task and a segmentation task. The first step of the reconstruction task realizes the implicit surface reconstruction through RGB-D observations to provide the scene geometry for the segmentation task. Secondly, the segmentation task builds a graph from the normal of mesh, and infers 3D pseudo IDs to associate the 2D instance IDs by instance mask of Grounded SAM. In addition, 3D instance ID corrects some of the erroneous semantic labels. Then, the second reconstruction step realizes 2D-3D labeling supervised by consistent semantic and instance labels, and finally obtains the panoptic mesh, point cloud, and novel view images of the scene.}
\vspace{-0.4cm}
\end{figure*}

In contrast to SAM which may provide over-segmented masks, Grounded-SAM~\cite{ren2024grounded} is a promising alternative to provide open-vocabulary semantic and instance labels of the desired granularity given a few text queries. But there are still two remaining challenges as shown in Fig~\ref{Fig_main}: First, language-conditioned instance segmentation VLM yields \textit{partial labeling}~\cite{fm-fusion} that only pixels corresponding to the class prompt have a valid prediction. In this way, the labeling result can be regarded as semantic segmentation with additional classes of unknown, which may cause lower accuracy for known classes. Moreover, only pixels with labels belonging to known classes can be supervised, thus insufficient for labeling the full 3D space. 
Second, \textit{instance association} across multiple frames is challenging. In~\cite{gaussian_grouping}, the association is derived by tracking~\cite{cheng2023tracking}, which fails for large frame in-between gap and long-term re-identification. In~\cite{Chen2024PVLFF, bhalgat2023contrastive}, contrastive learning is employed to build the embedding space of the instance, avoiding the association. When clustering the embedding space for instance segmentation, it is hard to tune the coefficients e.g. instance count. Panoptic Lifting~\cite{lifting} is proposed to infer the instance association using linear assignment, which may lose the global uniqueness due to the initial randomization sensitivity.

In this work, we propose a novel zero-shot panoptic reconstruction method, named PanopticRecon, by combining open vocabulary instance segmentation, Grounded-SAM~\cite{ren2024grounded}, and neural implicit surface reconstruction~\cite{nfatlas}. We tackle both challenges mentioned above, by 1) propagating partial labels to the full scene via a generalizable network and 2) building a 3D instance graph from 2D instance masks to establish globally consistent association. Specifically, to address the challenge of partial labeling, we propose to distill the \textit{dense} semantic VLM features to the 3D space, and learn a generalizable point-level classifier supervised by the \textit{partial} labels, effectively enabling label propagation. For the second challenge of instance association, we first reconstruct the scene geometry without labeling, and then leverage the reconstruction and 2D instance-level segmentation to build pseudo instance GTs in the 3D space to guarantee the ID uniqueness globally. The pseudo GT is acquired by graph inference, which is stated as the edge cutting to maximize the 3D-2D instance mask consistency. Assembling the improved prediction and association into the neural implicit representation, PanopticRecon arrives at zero-shot panoptic surface reconstruction without human annotation, which achieves superior accuracy in geometry and semantics as demonstrated in experiments. In summary, the contributions involve:
\begin{itemize}
    \item We propose a novel neural implicit representation-based method to leverage Grounded-SAM for zero-shot 3D dense panoptic reconstruction.
    
    \item We propagate partial labels with the aid of DINOv2, generate instance association by instance graph inference and correct semantic labels based on unique instance IDs, to address the limitations of Grounded-SAM results.
    
    \item Both indoor and outdoor sequence experiments to evaluate the effectiveness of outcomes including panoptic segmentation and mesh geometric quality.

\end{itemize}

%% file: sec2_related.tex
\section{Related Works}

\subsection{Semantic Neural Fields}
Initially, NeRF~\cite{mildenhall2021nerf} modeled only the radiance and geometry of 3D scenes, and some subsequent studies have used neural implicit representations to extend semantic information to the tasks of 3D scene modeling. Semantic NeRF~\cite{semanticnerf} first explored encoding semantics into NeRF by adding a separate branch to predict semantic labels, fusing noisy 2D semantic segmentation into a consistent volumetric model. NeSF~\cite{nesf} then predicts semantic fields by feeding density fields to a 3D semantic segmentation model. To achieve open-vocabulary semantic segmentation, NIVLFF~\cite{nivlff} combines NeRF and features from a foundation model, distilling visual features into an implicit representation model and achieving semantic segmentation of open-vocabulary scenes by mapping class texts embedding to distilled features.

\subsection{Panoptic Neural Fields}
Kirillov et al~\cite{panopticsegment} defined the 2D panoptic segmentation task and its metrics for the first time. In recent years, some works~\cite{pnf,panopticnerf,bhalgat2023contrastive,lifting} have adopted NeRF to solve the 3D scene panoptic segmentation problem. PNF~\cite{pnf} first proposed a method to obtain panoptic radiation fields of dynamic 3D scenes from images. The method takes the 3D bounding boxes of dynamic instances with known motion trajectories as the sampling space for each instance, with each instance having its own MLP. Panoptic NeRF~\cite{panopticnerf} also uses the 3D bounding box to assist panoptic segmentation which focuses more on the refinement of the 3D polygon labels, requiring more manual annotation in the early stage. To automatically obtain the 3D bounding boxes of objects to assist in supervising instances, Instance NeRF~\cite{instancenerf} proposes to determine the 3D bounding box of the object through the density reconstructed in advance, but it is only suitable for relatively simple scenes. Since 3D supervision is not easy to obtain, Panoptic Lifting~\cite{lifting} proposes to add linear assignments in the optimization process to align the instance IDs of multiple viewpoints, while losing the global uniqueness. Contrastive Lift~\cite{bhalgat2023contrastive} designs a contrastive formulation to enhance 2D segmentation to 3D by feature clustering. 

All the above methods apply a 2D pre-trained 2D segmentation model, which can only handle close-vocabulary scenes. For open-vocabulary scenes, PVLFF~\cite{Chen2024PVLFF} distills 2D features from two different VLMs for the semantic task and the instance task respectively, and realizes the mask label alignment in 3D by making the features in the SAM~\cite{sam} mask similar. However, two different foundation models lead to inconsistent labeling of semantics and instances, and the characteristics of SAM itself cause PVLFF to be unable to segment object-level instances accurately.

%% file: sec3_method.tex
\section{Method}

% \begin{figure*}[!htbp]
% \centering
% \includegraphics[width=\linewidth,keepaspectratio]{fig/pipeline.jpg}\\
% \vspace{-0.3cm}
% \caption{\label{Fig_pipeline}Our panoptic reconstruction system consists of a reconstruction task and a segmentation task. The first step of the reconstruction task realizes the implicit surface reconstruction through RGB-D observations to provide the scene geometry for the segmentation task. Secondly, the segmentation task builds a graph from the normal of mesh, and infers 3D pseudo IDs to associate the 2D instance IDs by instance mask of Grounded SAM. In addition, 3D instance ID corrects some of the erroneous semantic labels. Then, the second reconstruction step realizes 2D-3D labeling supervised by consistent semantic and instance labels, and finally obtains the panoptic mesh, point cloud, and novel view images of the scene.}
% \vspace{-0.5cm}
% \end{figure*}

\subsection{Overview}

Given a set of posed multi-view images and depth of a scene, the proposed zero-shot panoptic reconstruction can construct an implicit representation that encodes the SDF, color, semantics, and instance, with the aid of Grounded-SAM. The pipeline is fully automatic as shown in Fig.~\ref{Fig_pipeline}, of which the only information provided by the human user is the class text. We divide the pipeline into two main components, reconstruction and segmentation. In segmentation, we employ Grounded-SAM to label the instance with given classes and associate 2D instances by 3D mesh based graph inference, resulting in zero-shot multi-view images with semantic and consistent instance labels for panoptic reconstruction. In reconstruction, we have two stages of reconstruction. In the 1st stage, we build an initial 3D pure geometric map to help instance association. In the 2nd stage, we build a panoptic map by jointly learning SDF, color, semantics, and instances, using labeled images and DINOv2 as supervision, yielding the final panoptic segmented mesh and novel view images. 
% \vspace{-0.2cm}

\subsection{Zero-shot Panoptic Label Preparation}

We first identify a set of interested classes as text prompts for Grounded-SAM to label images with instances belonging to these classes. When pixels are labeled with more than one instance, we assign them the label with the highest score. We group the instances with the same class into semantic masks and divide all semantic classes into \textit{thing} and \textit{stuff} categories. Note that the per-frame instances are not associated, the semantic masks may be incomplete and Grounded-SAM may miss or misclassify instances in a single frame, which are the main issues to address for panoptic reconstruction.
% \vspace{-0.2cm}

\subsection{Instance Association and Label Correction}
\textbf{Graph Construction:}
We begin by building an initial 3D pure geometric map for instance association. The technical detail for the implicit surface reconstruction is shown in~\ref{section:E}. Given the 3D map and the 2D instance labeled images, we state the association problem as a graph inference problem.  %3D instance segmentation problem into a graph segmentation problem to achieve the association problem with minimum complexity $O$. 
First, we apply~\cite{felzenszwalb2004efficient} to compute the similarity between mesh vertices based on their normal directions, upon which the map is clustered into superfaces to reduce complexity. The center point of each superface is called a superpoint, which are modeled the nodes $\mathcal{V}$ of the scene graph $G=(\mathcal{V},\mathcal{E})$. Initialize that edges $\mathcal{E}$ exist between all nodes and that each edge has zero vote count. The final vote counts determine whether or not to cut off the edges.
 
%Unlike~\cite{SAM-guided Graph Cut}, our 2D masks are instance-level segmentation masks, instead of the different levels of segmentation masks generated by SAM~\cite{} based on the projection point prompt. We have the advantage of instance segmentation in 2D without the need to implicitly encode nodes and edges by the IoU of the multi-level masks prompted by different node projections. 

For an image $I$, we define the set of nodes with valid projection masks as $\mathcal{V}_I \subset \mathcal{V}$. Then we define the overlap $U_{ij}$ as the percentage of the intersection area between the projection mask of the superface corresponding to the node $V_i\in \mathcal{V}_I$, and the 2D instance mask $M_j$ in the image $I$. For the set of nodes $\mathcal{V}_{I}^j \subset \mathcal{V}_I$ with non-zero $U_{ij}$, we define the center node of $M_j$:
\begin{equation}
    V_c^j = \arg\max_{V_i \in \mathcal{V}_{I}^j} U_{ij}
\end{equation}
For remaining nodes $V_k \in \mathcal{V}_{I}^j$, if the node has $U_{kj}$ greater than a threshold, we vote the edge between $V_k$ and $V_c^j$. For nodes in $\mathcal{V}_I -\mathcal{V}_{I}^j$, we deduct a vote of their edges to $\mathcal{V}_{I}^j$. By iterating all images, we derive a resultant graph, in which we reserve only edges with positive vote counts.

%assign connectivity scores $X$ of edges$\{E\}$ by

%When  is regarded as part of the instance, the connection score of the edge between the superpoint and the center node of the instance mask is increased.  

\begin{figure}[!t]
\centering
\includegraphics[width=\linewidth,keepaspectratio]{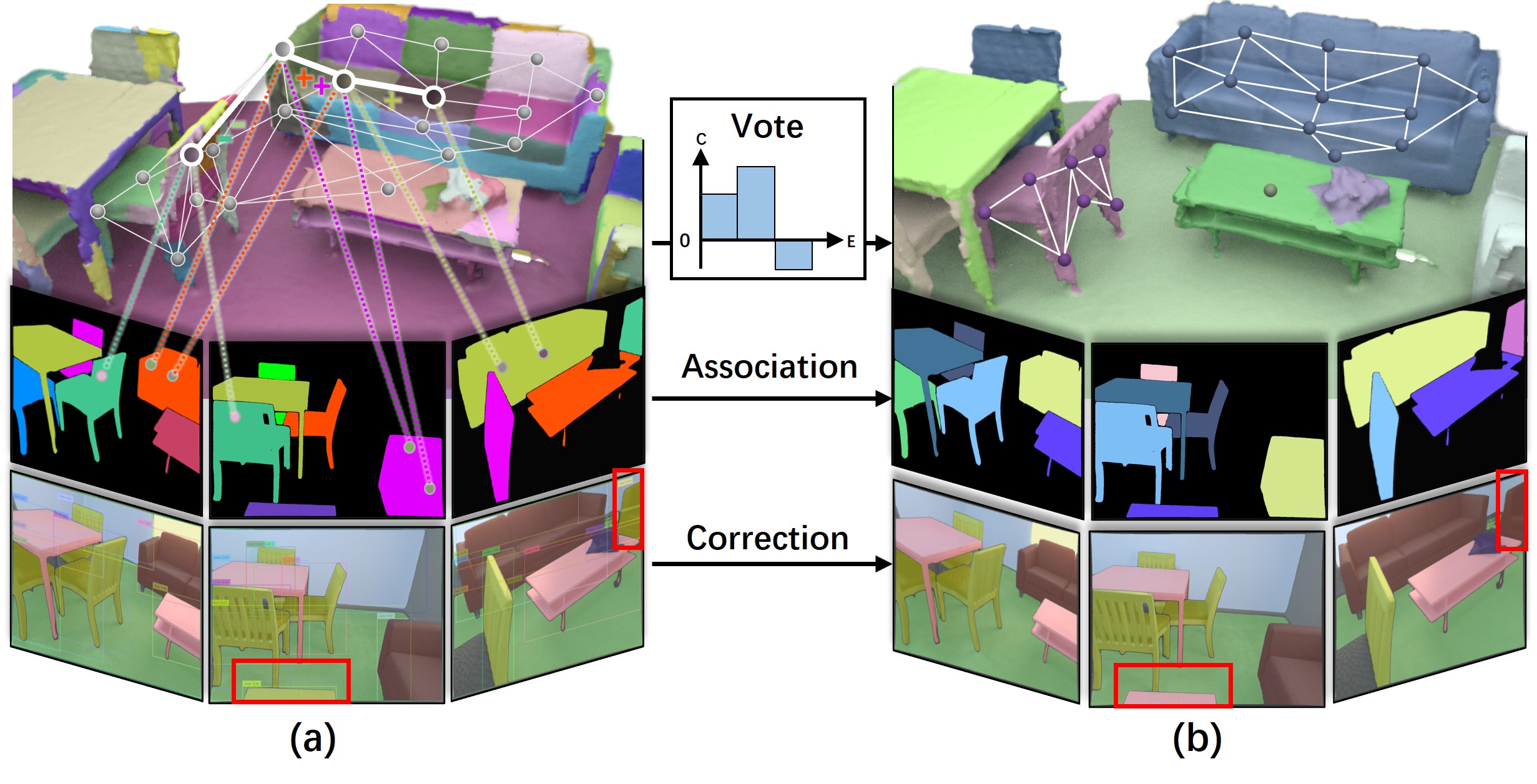}\\
\vspace{-0.4cm}
\caption{\label{Fig_graph} The points in the upper graph in \textbf{(a)} are the nodes (superpoints) of the graph. The color region corresponding to each node is the superface. We determine the nodes in an instance mask of a frame based on the overlap between the instance mask provided by Grounded SAM and the mask projected by the superface, and vote for the edges between the selected nodes. Similarly, we reduce the votes for the edges between nodes corresponding to masks of different instances in that frame. The edges with non-positive votes are finally cut and the nodes connected form an instance as shown in \textbf{(b)}. Once the 3D instance pseudo IDs are obtained, we associate 2D instance IDs while correcting incorrect semantic labels.}
\vspace{-0.5cm}
\end{figure}

\textbf{Association and Correction:}
As shown in Fig~\ref{Fig_graph}, on the resultant graph, we apply the graph clustering and regard the clusters as 3D instances. By assigning scene level globally unique IDs to 3D instances, we project them onto all 2D images and align them with the 2D instance by evaluating the IoUs with the projection masks, so that the 2D instance in each image can be associated with a unique 3D ID. Since 3D instances satisfy multi-view consistency, we determine their instance class by voting from labels of all associated 2D instance classes and furthermore correct the 2D labels to refine the semantic segmentation.

\subsection{Label Propagation}

Given the image with partial semantic labeling corrected after association, we further propagate the labels to the whole 3D scene. By integrating the semantic images in the 3D volume, a 3D point not labeled in one semantic image, may be compensated by other images, thus the 3D scene labeling is expected to be improved. The remaining problem is to propagate the labels to points that are unknown in all images. 

\textbf{DINOv2 Aided Semantics:} 
The propagation is achieved by utilizing the image cues. Specifically, given an RGB image $I$, its corresponding DINOv2 feature is denoted as $V$. We reduce the feature dimension of $V$ by PCA to 64-dimensional compressed features $V_{64}$. Using these features as the supervision, we distill them to the intermediate layers of the semantic branch. As shown in Fig.~\ref{Fig_network}, concatenating the distilled feature with the 3D point position encoding, the semantic labeling of this point is actually stated as a per-point classification problem. The classifier is MLP, and the input is the feature co-supervised by DINOv2 and semantic labels. Since DINOv2 feature is believed to be semantic meaningful, the generalization to unknown points is expected.

\begin{figure}[!t]
\centering
\includegraphics[width=\linewidth,keepaspectratio]{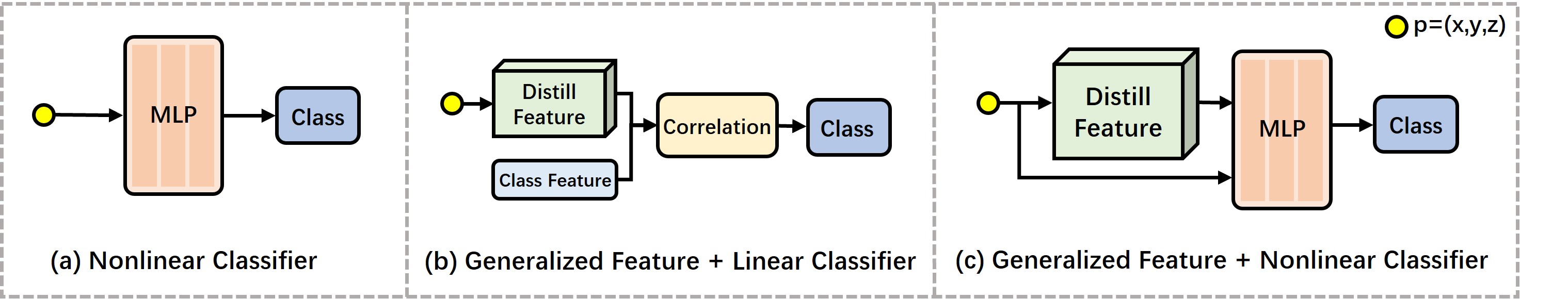}\\
\vspace{-0.3cm}
\caption{\label{Fig_segmentation} The flow of the segmentation of the previous work is shown as \textbf{(a)} and \textbf{(b)}, and ours is shown as \textbf{(c)}.}
\vspace{-0.4cm}
\end{figure}

\textbf{Discussion on Classifier:} 
In contrast, existing open-vocabulary semantic segmentation methods~\cite{Peng2023OpenScene,Chen2024PVLFF,nivlff} predict the labels by correlating the class language feature with distilled VLM features. As shown in Fig.~\ref{Fig_segmentation}, in the perspective of per-point classification, these methods employ a linear classifier for labeling. Considering the nonlinear 2D semantic decoder is usually required after the original 2D image VLM feature~\cite{lseg, oquab2023dinov2}, such linear classification of the 3D point may be too simple for an accurate result even with distilled VLM features.

% \vspace{-0.2cm}
\subsection{Neural Implicit Representation}
\label{section:E}

\begin{figure}[!t]
\centering
\includegraphics[width=\linewidth,keepaspectratio]{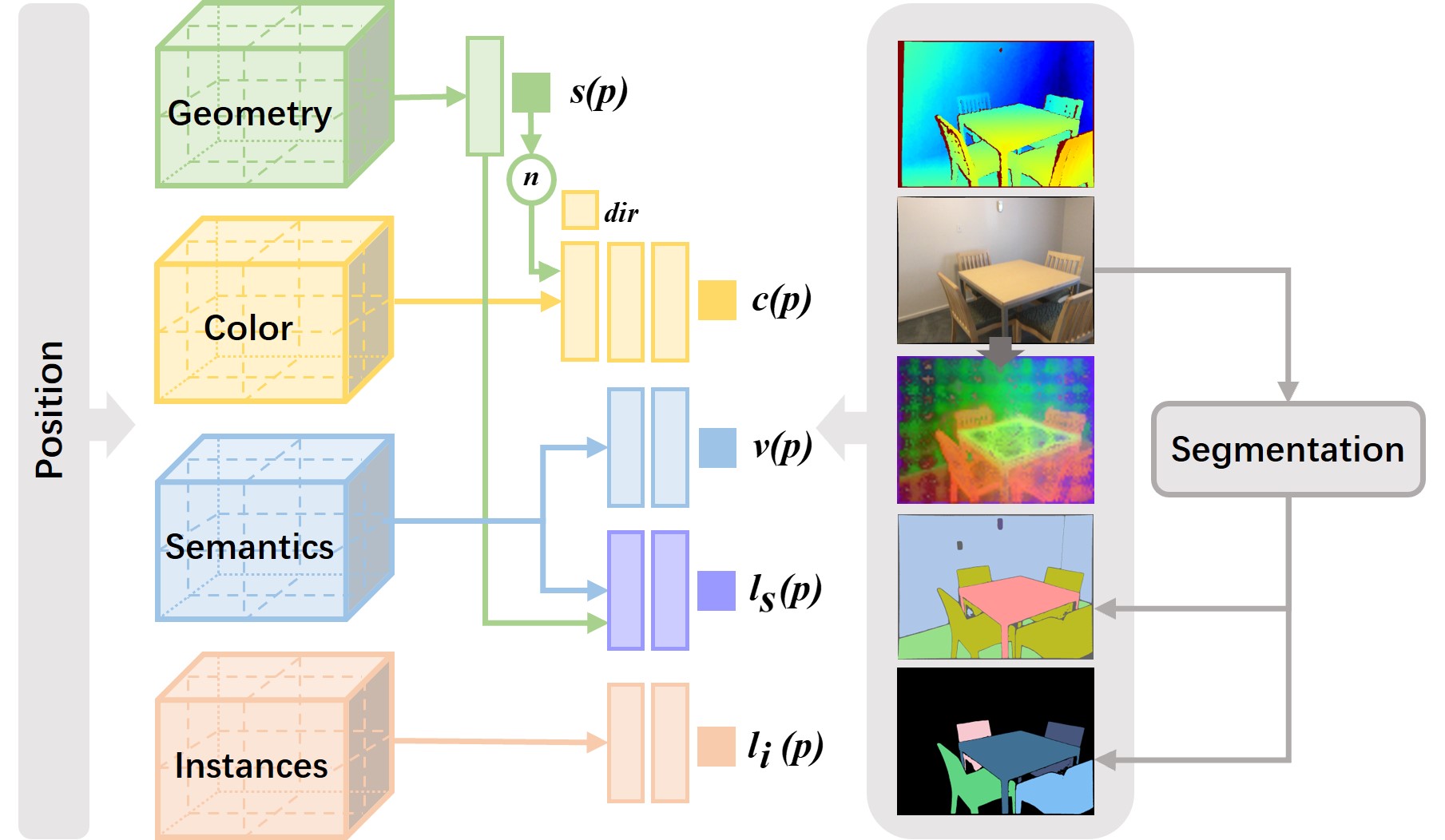}\\
\vspace{-0.3cm}
\caption{\label{Fig_network} The network architecture which is described in detail in Sec.E.}
\vspace{-0.4cm}
\end{figure}

With the corrected labels, instance association, and the DINOv2 aided semantic label propagation, we derive the panoptic reconstruction of the scene by learning an implicit neural representation. As shown in Fig.~\ref{Fig_network}, the network of PanopticRecon consists of hierarchical hash encoding~\cite{muller2022instant} and multitask branches. We design independent feature volumes for geometry, color, semantics, and instances and optimize the features jointly, where the task branches interact with each other to complement the accuracy. 

\textbf{Feature Volumes:}
As shown in Fig.~\ref{Fig_network}, our representation consists of 4 feature volumes including geometry, color, semantics and instance, denoted as $\Psi_{c}$, $c\in\{G,C,S,I\}$. Given a 3D point, the feature is yielded by querying the corresponding volume. All these features are learnable by the supervision of multiple tasks.

\textbf{Volume Rendering:}
The representation is supervised at the ray level, thus we introduce the volume rendering to generate the prediction of the pixel corresponding to the ray. Based on the camera pose, we define the origin and direction of a ray $(o,d)$ passing through a pixel $x$ in image, depth, semantic or instance segmentation image. Along the ray, we sample $N$ successive 3D spatial points $\{p_i\}$ as $p_i = o + \rho_i d$, where $\rho_i$ is the distance. Based on the feature volumes and the downstream network processing as shown in Fig.~\ref{Fig_network}, each point is assigned with a prediction $f(p_i)$, which can be SDF $s$, color $c$, semantic probability $l_{s}$, distilled DINOv2 feature $v$, instance probability $l_{i}$, and depth $d$. We then have the volume rendering as:
\begin{equation}
    u_f(x) = \sum_{i=1}^{N}{T_{i} \alpha_{i} f(p_{i})}
    \label{vr}
\end{equation}
where $T_{i}=\prod_{m=1}^{i-1}(1-\alpha_{m})$, $\alpha_{m}$ is the discrete opacity value defined under the S-density function assumption~\cite{wang2021neus} as:
\begin{equation}
    \alpha_{i} = \max\left(\frac{ \Phi(s_i)-\Phi(s_{i+1}) }{ \Phi(s_i) }, 0\right) 
\end{equation}
where $s$ is SDF of the point, $\Phi(s)$ is Sigmoid function $\Phi(s)=(1+e^{-\xi s})^{-1}$ with a temperature coefficient $\xi$.

\textbf{Network Architecture:} Given a 3D point $p_i$, the geometry feature volume maps $p_i$ to a feature vector $\Psi_{G}(p_i)$. A small MLP $h$ is applied to yield the SDF value:
\begin{equation}
    s(p_i) = h_{s}(\Psi_{G}(p_i))
\end{equation}
Its normal is derived as:
\begin{equation}
    n(p_i) = \frac{\partial s}{\partial p_i}
\end{equation}
Then the color is derived by querying the color volume feature $\Psi_{C}(p_i)$, the direction encoding $\lambda(p_i)$, and the normal $n(p_i)$, which are passed through a small MLP $h_c$ as:
\begin{equation}
    c(p_i) = h_{c}(\Psi_{C}(p_i), \lambda(p_i), n(p_i))
\end{equation}
Together with the semantic label probability $l_s(p_i)$, distilled DINOv2 $v(p_i)$ is also generated by querying the semantics volume feature $\Psi_{S}(p_i)$ as well as geometric volume $\Psi_{G}(p_i)$, which are passed through a small MLP $h_s$ as:
\begin{equation}
    l_s(p_i), v(p_i)= h_{s}(\Psi_{S}(p_i), \Psi_{G}(p_i))
\end{equation}
Finally, the instance label probability $l_i(p_i)$ is generated by querying the instance volume feature $\Psi_{I}(p_i)$, which is passed through a small MLP $h_i$ as:
\begin{equation}
    l_i(p_i)= h_{i}(\Psi_{I}(p_i))
\end{equation}

\begin{table*}[t]
    \centering
    \setlength{\tabcolsep}{4pt}
    \vspace{-0.3cm}
    \caption{Panoptic Segmentation quality using different methods on indoor and outdoor datasets}
    \centering
    \vspace{-0.3cm}
    % \resizebox{\linewidth}{!}{
    \begin{tabular}{lccc|cc|cc|ccc|cc|cc}
        \toprule
            \multirow{2}*{\textbf{Method}} & \multicolumn{7}{c}{ScanNet(V2)}& \multicolumn{7}{c}{KITTI-360}\\
            \cmidrule(r){2-8} \cmidrule(r){9-15}
             & $\text{PQ}^{\text{s}}$$\uparrow$ & SQ$\uparrow$ & RQ$\uparrow$ %& mIoU$\uparrow$ 
             & mIoU$\uparrow$ & mAcc$\uparrow$ 
             & mCov$\uparrow$ & mW-Cov$\uparrow$ &
             $\text{PQ}^{\text{s}}$$\uparrow$ & SQ$\uparrow$ & RQ$\uparrow$ %& mIoU$\uparrow$ 
             & mIoU$\uparrow$ & mAcc$\uparrow$ 
             & mCov$\uparrow$ & mW-Cov$\uparrow$ \\
            
        \midrule
        \textit{Close-vocabulary:} \\
            GT + Panoptic NeRF & 52.12 & 59.50 & 71.35 %& 72.42 %pan
            & 52.65 & 61.28 %sem
            & 61.79 & 72.04 %ins
            & 42.61 & 51.38 & 53.01 %& 83.80 %pan
            & 46.90 & 59.73 %sem
            & 57.91 & 83.50 \\ 
        \midrule
        \textit{Open-vocabulary:} \\
            GS + Panoptic Lifting & 47.96 & 52.11 & 64.79 %& 71.85 %pan
            & 61.79 & 71.76 %sem
            & 46.61 & 64.77 %ins
            & 20.86 & 20.86 & 27.27 %& 76.50 %pan
            & 41.10 & 49.29 %sem
            & 6.63 & 14.64 \\ %ins
            PVLFF & 25.14 & 36.08 & 34.55 %& 67.29 %pan
            & 51.84 & 58.97 %sem
            & 44.94 & 48.15 %ins
            & 27.46 & 32.61 & 40.25 %& 63.08 %pan
            & 40.94 & 46.20 %sem
            & 12.22 & 21.34 \\ %ins
            \textbf{Ours} & \textbf{62.39} & \textbf{64.25} & \textbf{74.69} %& \textbf{79.59} %pan
            & \textbf{70.19} & \textbf{84.79} %sem
            & \textbf{65.67} & \textbf{77.87} %ins
            & \textbf{35.95} & \textbf{46.43} & \textbf{45.54} %& \textbf{80.10} %pan
            & \textbf{43.08} & \textbf{62.65} %sem
            & \textbf{36.80} & \textbf{57.08} \\ %ins
        \bottomrule
    \end{tabular}
    % }
    \label{Tab_panoptic}
    \vspace{-0.4cm}
\end{table*}

\textbf{Losses:} With all point-level predictions, we can generate pixel-level predictions by (\ref{vr}). We set up geometric loss, photometric loss, semantic loss, and instance loss to supervise the panoptic reconstruction.

The geometric loss consists of three parts:
\begin{equation}
    L_{SDF}(p_i) =\begin{cases}
                \vert s(p_i) - b(p_i)\vert  & |b|\leq \tau\\
                \max(0, e^{-\beta s(p_i)}-1, s(p_i) - b(p_i)) & o.w.
                \end{cases}
\end{equation}
where $\tau$ is a threshold to truncate SDF, $b(p_i)=D(x)-\rho_i$ is the distance between the $p_i$ and the observed depth of pixel $x$ along the ray as an approximated SDF. In addition, eikonal loss is used as regularization:
\begin{equation}
    L_{eik}(p_i) = \|1-|n(p_i)|\|^2
\end{equation}
Depth is also supervised at the pixel level for depth generated by volume rendering:
\begin{equation}
    L_{Depth}(x) = \vert D(x) - u_d(x)\vert
\end{equation}

The photometric loss is built between the rendered color and the ground-truth color: 
\begin{equation}
    L_{Color}(x) = {{\Vert I(x) - u_c(x)\Vert}}^2
\end{equation}

The semantic loss and instance loss share the same form. Each ray yields a semantic and instance probability distribution 
%$\hat{\Gamma_r^s}=\sum_{i=1}^{N}{T_{i} \alpha_{i} Softmax(l_s(p_{i})})$ over the semantic class $K=\{k\}$ and a instance probability distribution $\hat{\Gamma_r^i}=\sum_{i=1}^{N}{T_{i} \alpha_{i} Softmax(l_i(p_{i})})$ over the instance ID $J=\{j\}$, 
which are supervised by cross-entropy of $u_{l_s,l_i}$ with respect to instance and semantic image labels $I_s$ and $I_i$ as pseudo ground truth $\Gamma_r$:
\begin{equation}
    L_{S/I}(x) = - {\sum_{k\in K_{s,i}}{I_{s,i}(x) \log u_{l_{s,i}}(x)}} 
\end{equation}
where $K_s$ is the total number of semantic classes and $K_i$ instances. In addition, the DINOv2 loss for label propagation is built between rendering of distilled DINOv2 and the compressed features $V_{64}$ as: 
\begin{equation}
    L_{DN}(x) = {{\Vert V_{64}(x) - u_v(x)\Vert}}^2
\end{equation}

In a nutshell, our total loss $L$ is:
\begin{equation}
    L = L_{Depth}+L_{SDF}+L_{eik}+L_{Color}+L_{S/I}+L_{DN}
\end{equation}

\textbf{Two-stage Reconstruction:} The neural representation learning is conducted for two rounds. At the first time, no semantic and instance supervision is applied since there is no instance association and the corrected labels i.e. no $L_{S}$, $L_{dino}$ and $L_{I}$. But this first-round reconstruction helps the instance association and label correction as shown above. After that, given all supervision available, we have the second-round reconstruction, which finally yields the panoptic reconstruction of the scene.

%% file: sec4_exp.tex
\section{Experimental Results}

We validate the effectiveness of our method and compare with others in both indoor scenes and outdoor scenes to demonstrate the panoptic reconstruction performance. We also conduct an ablation study on the network components. 

%-----------------------------------
\vspace{-0.2cm}
\subsection{Setup}
\textbf{Dataset:}  Our study involves experiments on an indoor datasets: Scannet V2 \cite{dai2017scannet} and an outdoor dataset: KITTI-360 \cite{liao2022kitti}. ScanNet is an RGBD dataset that contains a large number of real indoor scenes with perspective-rich images captured by iPad Air2 and reconstructed geometry\cite{dai2017bundlefusion} as ground truth, which makes it suitable for evaluation tests. We use 3 scenes for our evaluation. KITTI-360 is a large-scale driving dataset with 2D/3D manually annotated ground truth and multimodal observations captured using a Velodyne HDL-64 LiDAR and 4 cameras. We select a part of seq00 with a length of $200m$ for evaluation. We use depth observations and one perspective camera in our experiments. 

\textbf{Metrics:}  We evaluate the panoptic reconstruction quality using scene-level panoptic quality ($PQ^{s}$) proposed by\cite{lifting}, which merges the segments belonging to the same instance identifier in all frames of a scene and compute the panoptic quality of the merged segments for evaluation. We also evaluated the $mIoU$, $mAcc$, $mCov$ and $mWCov$ for 2D semantic segmentation and instance segmentation, respectively, to show the effectiveness of different branches.

\textbf{Implementation:} 
We set the text prompt of Grounded-SAM by one or more related words per class. Since some classes are hard to discriminate even by human, we follow~\cite{lifting} to merge some classes into one class, such as \textit{sofa} and \textit{armchair}, \textit{ground} and \textit{sidewalk}, \textit{vegetation} and \textit{terrain}. For network architecture, the decoders for the geometric, semantic, and instance are all one-layer MLP of 128 hidden units, and the decoders for the color and DINOv2 branches are all 3-layer MLPs of 128 hidden units. All experiments run on a single A100 GPU. 

\textbf{Baselines:} We adopt the Panoptic NeRF~\cite{panopticnerf} as close-vocabulary model aided method, Grounded-SAM (GS) + Panoptic Lifting~\cite{ren2024grounded, panopticnerf}, and PVLFF~\cite{Chen2024PVLFF} methods as open-vocabulary model aided methods. The three state-of-the-art methods are capable of panoptic reconstruction using RGBD inputs. Regarding Panoptic NeRF, we provide the ground-truth 3D instance bounding box (GT), making it a reference method. The original Panoptic Lifting employs the close-vocabulary panoptic segmentation method, we replace it with results of Grounded-SAM, making it a comparative open-vocabulary model aided method.

\subsection{Comparative Study}
%--------------------------------------------------------
\begin{figure*}[!htbp]
\centering
\includegraphics[width=\linewidth,keepaspectratio]{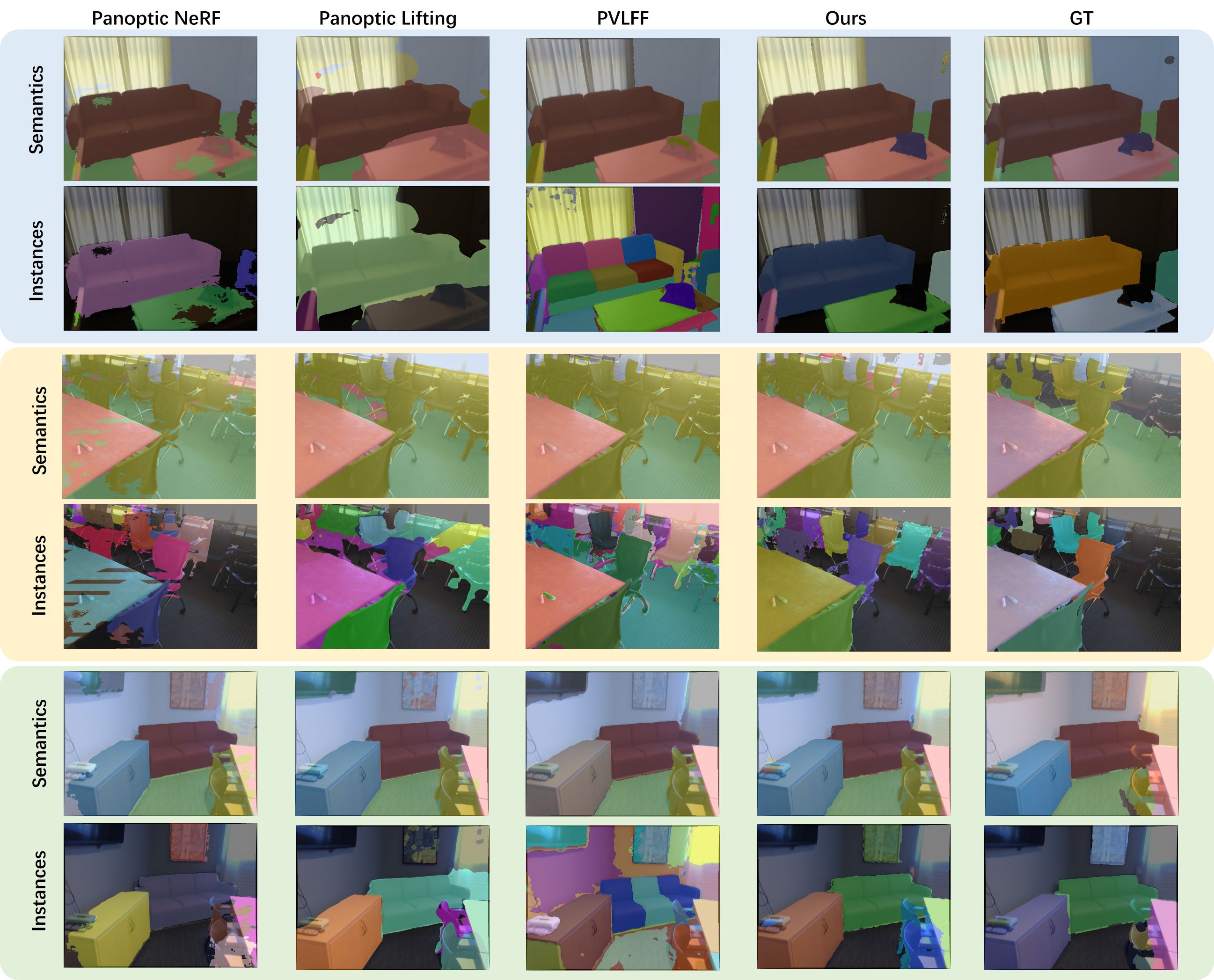}\\
\vspace{-0.3cm}
\caption{\label{Fig_render} Comparison of the quality of semantic segmentation and instance segmentation of different methods on ScanNet.}
\vspace{-0.5cm}
\end{figure*}

\begin{figure*}[!htbp]
\centering
\includegraphics[width=\linewidth,keepaspectratio]{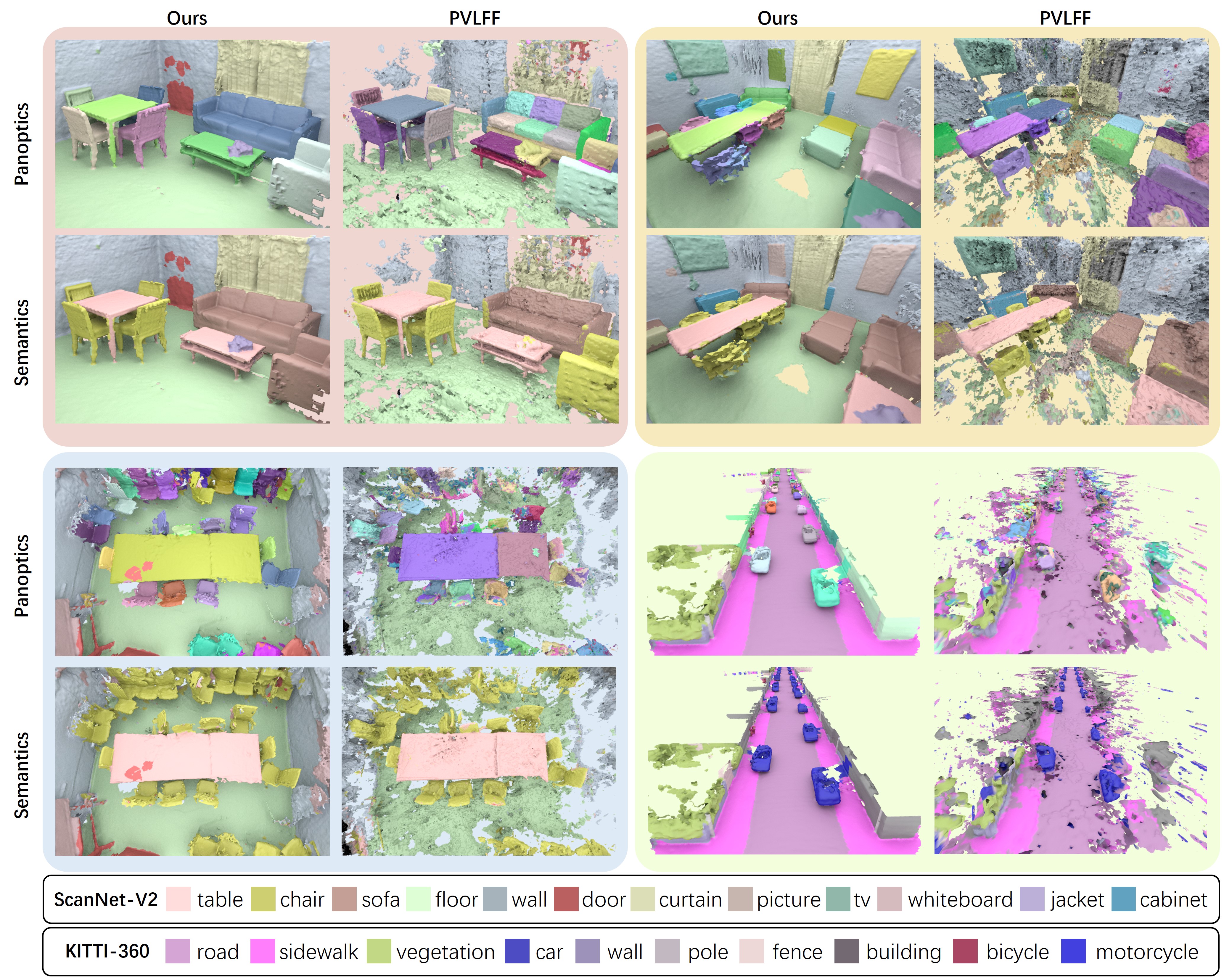}\\
\vspace{-0.3cm}
\caption{\label{Fig_mesh} Comparison of the quality of panoptic reconstruction and semantic reconstruction on ScanNet and KITTI-360 by different methods. }
\vspace{-0.3cm}
\end{figure*}

\textbf{Semantic Segmentation:} 
We first compare the semantic segmentation. Panoptic Nerf uses close-vocabulary model PSPNet~\cite{pspnet} to provide 2D semantic images. For KITTI-360, we use the segmentation model pre-trained on CityScape~\cite{cordts2016cityscapes} as that in~\cite{panopticnerf}. For ScanNet, as there is no reference model, we use the segmentation model pre-trained on ADE20k~\cite{ade20k} that is also collected in indoor environment. As shown in Tab.~\ref{Tab_panoptic}, our semantic segmentation metrics are comparable to those of Panoptic Nerf on KITTI-360, and superior to those on ScanNet, reflecting the limitation of close-vocabulary semantic segmentation performance on novel dataset that even collected from similar scene. Compared with other open-vocabulary segmentation based methods, our method shows better performance, validating the effectiveness of DINOv2 to propagate the partial labels to the whole image.

\textbf{Instance Segmentation:} 
We then compare the instance segmentation in Tab.~\ref{Tab_panoptic}. As Panoptic Nerf employs GT 3D bounding boxes, its performance is unsurprisingly far better than the others on KITTI-360. In ScanNet, the bounding boxes of different instances overlap with each other more frequently than in KITTI-360, degenerating the instance segmentation performance of Panoptic Nerf. The error in semantic segmentation further affects the extraction of instance masks. In contrast, thanks to the explicit graph inference to associate the instances and correct the class labels, our method achieves better performance. Compared with other open-vocabulary methods, our method also outperforms. Note that the non-uniqueness causes inferior performance of Panoptic Lifting, especially on KITTI-360, which has a large number of instances. For the PVLFF, it is hard to correctly cluster instances with hand-crafted coefficients and rules.

\textbf{Panoptic Segmentation:} 
We finally compare the panoptic segmentation in Tab.~\ref{Tab_panoptic}. On KITTI-360, Panoptic Nerf is naturally the best method with the aid of both ground truth bbox and semantic segmentation model trained on a similar dataset. Our method is slightly inferior due to the instance segmentation. For ScanNet, with the aid of open-vocabulary model and proposed techniques, our method performs best, which shows its zero-shot capability when close-vocabulary model are constrained by the generalization gap. 

\begin{table}[t]
    \centering
    \setlength{\tabcolsep}{4pt}
    % \vspace{-0.2cm}
    \caption{Mapping quality using different methods on ScanNet}
    \centering
    \vspace{-0.2cm}
    % \resizebox{\linewidth}{!}{
    \begin{tabular}{lccc}
        \toprule
            \textbf{Method} & Comp (cm)$\downarrow$ & Acc (cm)$\downarrow$ & C-L1 (cm)$\downarrow$\\
        
        \midrule
            PVLFF  & 4.55 & 14.04 & 9.30  \\
            Ours & \textbf{1.84} & \textbf{1.75} & \textbf{1.80}\\
        \bottomrule
    \end{tabular}
    % }
    \label{Tab_mesh}
    \vspace{-0.5cm}
\end{table}

Compared with other open-vocabulary model aided methods, the benefits of our method are discussed in the semantic and instance segmentation above. Addressing instance association and partial labeling overcomes the limitations of an open-vocabulary segmentation model, leading to improved performance in zero-shot panoptic surface reconstruction.

%in performances are shown in Tab.~\ref{Tab1}. Based on the semantic and instance segmentation performance,  shows the evaluation results of our method and the baselines. ...
%Our approach is compared against three SOTA methods, Panoptic NeRF~\cite{}, Panoptic Lifting~\cite{} and PVLFF~\cite{}, on ScanNet V2 and KITTI-360 for evaluation. 

\textbf{Mesh Reconstruction:}
To evaluate the reconstruction accuracy, we compare the surface accuracy with PVLFF which also employs depth supervision. As shown in Tab.~\ref{Tab_mesh}, our method demonstrates the best performance, which is explained by the SDF formulation, depth observations, as well as better labeling. As several cases are shown in Fig.~\ref{Fig_mesh}, our method can recover better surface details.

\subsection{Ablation Study}

We then present the ablation study for components to show its contribution to the whole pipeline, including label propagation, instance association, and label correction.

\textbf{Effect of Label Propagation:} 
To evaluate the label propagation, we compare our network with and with the aid of DINOv2. As shown in Tab.~\ref{Tab_ablation}, with the label propagation, the partial labeling commonly occurring in open-vocabulary segmentation is relieved.

\begin{table}[t]
    \centering
    \setlength{\tabcolsep}{4pt}
    % \vspace{-0.2cm}
    \caption{Ablation to the semantic segmentation quality on ScanNet}
    \centering
    \vspace{-0.2cm}
    % \resizebox{\linewidth}{!}{
    \begin{tabular}{cccc}
        \toprule
            DINOv2 & Correction & mIoU$\uparrow$ & mAcc$\uparrow$  \\
        \midrule
            \ding{55} & \ding{55} & 66.10 & 79.25 \\
            \ding{55} & \ding{51} & 67.24 & 80.53  \\
            \ding{51} & \ding{55} & 68.43 & 83.08 \\
            \ding{51} & \ding{51} & \textbf{70.19} & \textbf{84.79}\\
        \bottomrule
    \end{tabular}
    % }
    \label{Tab_ablation}
    \vspace{-0.5cm}
\end{table}

%--------------------------------------------------
\textbf{Effect of Instance Association:}
To evaluate the instance association, we replace the graph inference with two existing options: label tracking~\cite{gaussian_grouping} and linear assignment~\cite{lifting}. To track the label across frames, we employ DAVA~\cite{cheng2023tracking}. The results are shown in Tab.~\ref{Tab_association}, which tells the advantage of an explicit graph inference on the global scene: uniqueness and multi-view consistency, compared with linear assignment and local tracking. Note that the linear assignment performs worse than Panoptic Lifting due to the lower randomness of the initial ID distribution when using a feature volume than a large MLP.

\begin{table}[t]
    \centering
    \setlength{\tabcolsep}{4pt}
    \vspace{-0.3cm}
    \caption{Ablation of instance association methods to the quality of instance segmentation on ScanNet}
    \centering
    \vspace{-0.2cm}
    % \resizebox{\linewidth}{!}{
    \begin{tabular}{lcccl}
        \toprule
            \textbf{Method} & mCov$\uparrow$ & mW-Cov$\uparrow$  \\
        
        \midrule
            Tracking (DAVA)  & 55.22 & 59.93  \\
            Linear Assignment  & 31.81 & 50.93  \\
            Graph Segmentation (Ours) & \textbf{65.67} & \textbf{77.87}\\
        \bottomrule
    \end{tabular}
    % }
    \label{Tab_association}
    \vspace{-0.5cm}
\end{table}

%--------------------------------------------------
\textbf{Effect of Label Correction:} To evaluate the contribution of label correction, we compare with the method using labels without correction. Since prediction error is unavoidable in open-vocabulary methods, an instance may be assigned with different classes across frames, which divides the 3D instance into different labels, resulting in weaker performance.

%% file: sec5_conclusion.tex
% \vspace{-0.2cm}
\section{Conclusion}
% \vspace{-0.1cm}
In this paper, we propose to endow the neural surface reconstruction with the open-vocabulary segmentation method Grounded-SAM, enabling the zero-shot panoptic reconstruction. As a cost, we have to address the partial labeling and instance association. Accordingly, we propose graph inference, label correction and label propagation, and build a neural implicit representation based panoptic reconstruction pipeline. The results demonstrate the effectiveness of PanopticRecon over existing zero-shot reconstruction methods and close-vocabulary prediction aided reconstruction methods.

%% file: bibliography.tex
%\bibliographystyle{gbt7714-2005}
%\bibliography{bibliography}
\bibliographystyle{IEEEtran}
\bibliography{IEEEabrv,bibliography}

% \addcontentsline{toc}{chapter}{\bibname}